# SAM*: Task-Adaptive SAM with Physics-Guided Rewards

Kamyar Barakati[1,*], Utkarsh Pratiush[1], Sheryl L. Sanchez[1], Aditya Raghavan[1], Delia J. Milliron[2], Mahshid Ahmadi[1], Philip D. Rack[1], and Sergei V. Kalinin[1,3, †]

[1] *Department of Materials Science and Engineering, University of Tennessee, Knoxville, TN 37996 USA*
[2] *Department of Chemical Engineering, University of Michigan Ann Arbor, MI 48109 USA*
[3] *Pacific Northwest National Laboratory, Richland, WA 993543 USA*

**Abstract**

Image segmentation is a critical task in microscopy, essential for accurately analyzing and interpreting complex visual data. This task can be performed using custom models trained on domain-specific datasets, transfer learning from pre-trained models, or foundational models that offer broad applicability. However, foundational models often present a considerable number of non-transparent tuning parameters that require extensive manual optimization, limiting their usability for real-time streaming data analysis. Here, we introduce a reward function-based optimization to fine-tune foundational models and illustrate this approach for SAM (Segment Anything Model) framework by Meta. The reward functions can be constructed to represent the physics of the imaged system, including particle size distributions, geometries, and other criteria. By integrating a reward-driven optimization framework, we enhance SAM's adaptability and performance, leading to an optimized variant, SAM*, that better aligns with the requirements of diverse segmentation tasks and particularly allows for real-time streaming data segmentation. We demonstrate the effectiveness of this approach in microscopy imaging, where precise segmentation is crucial for analyzing cellular structures, material interfaces, and nanoscale features.

---

[*] K.barakat@vols.utk.edu
[†] Sergei2@utk.edu

**Introduction**

Modern imaging produces vast volumes of heterogeneous data at high rates. Turning these data into measurements, decisions, and control signals requires reliable partitioning of scenes into meaningful units. With applications from clinical planning[1] and autonomous driving[2] to microstructure analysis in microscopy[3], image segmentation is the gateway from pixels to reproducible science and engineering.[1, 4]

Historically, image analysis has been a post-processing step.[5] Experiments were run, data was captured and archived, and segmentation was performed later. This allowed for computationally intensive processing, broad parameter exploration, and expert review. This allowed for extensive computation and expert review, but at the cost of significant latency: results arrived long after acquisition was complete, leaving no opportunity to adapt the experiment in real-time. Consequently, subtle phenomena were missed, and suboptimal configurations went uncorrected.[6-9]

Classical segmentation drew on fixed image-processing and early machine-learning pipelines rather than learned representations. Common approaches included global or adaptive thresholding (Otsu)[10] to separate foreground and background; edge detection and region growing, with variants such as the watershed transform[11], to delineate touching objects; active contours that deform a curve under gradient and smoothness terms; unsupervised clustering in feature space (k-means[12], mean-shift[13]) to group similar pixels without labels; and morphological filtering combined with graph-based formulations (graph cuts[14], normalized cuts[15]) to impose global consistency. These methods are fast, interpretable, and effective when assumptions such as homogeneous regions, clear boundaries, stable illumination, and moderate noise hold. However, these methods require careful parameter tuning, are sensitive to acquisition conditions, and hence often fail to generalize across samples, instruments, or modalities. Graph formulations can be costly on large images, while clustering typically ignores spatial coherence without additional regularization. In low-contrast settings such as microscopy, variations in illumination, noise, or specimen preparation frequently defeat fixed thresholds and edge cues, prompting manual or semi-manual intervention.

Against this drawback, Deep convolutional neural networks emerged to address these limitations by learning pixel-wise predictors end to end. Early milestones, Fully Convolutional Networks[16] and U-Net[17] established encoder–decoder architectures with skip connections that recover fine spatial detail while leveraging deep semantic context. Subsequent Convolutional Neural Network (CNN) and transformer variants extended multi-scale reasoning (dilated convolutions, pyramid pooling)[18] and broadened the task to instance and panoptic segmentation. Across domains like light and electron microscopy, these models delineate cells, organelles, and microstructural phases more accurately and consistently than classical pipelines. In practice, pretrained backbones and released models often perform well when the target data resemble the training distribution, enabling fast, consistent inference. Performance, however, degrades under domain shift (modality, contrast, preparation) and typically requires additional data and fine-tuning.[19]

Recent advances in computer vision and automation are enabling a paradigm shift toward real-time image analysis. This transition is critical for two principal reasons: (i) it provides investigators with actionable feedback for in situ decision-making, and (ii) it enables fully automated experimental workflows. Integrated with dynamic acquisition and control, real-time analysis closes the loop, elevating data processing from a passive report to an active control primitive.

The key challenge is adapting our experimental workflows to a real-time paradigm. This transition requires managing end-to-end latency and developing algorithms that are robust to constantly changing experimental conditions like drift, sensor noise, and motion blur, all while minimizing human input. An obstacle is the traditional reliance on supervised training, which creates a data bottleneck due to the high cost and inconsistency of dense, pixel-wise labels. This makes it difficult for models to generalize to new data without extensive re-training.[20]

In microscopy, frameworks like AtomAI enables pixel wise segmentation, image to spectrum mapping, and ensemble-based uncertainty for quantitative analysis, but it depends on dense labels.[21, 22] ResHedNet[23] yields precise grain boundary and domain wall delineation with ensemble based reliability, yet it needs edge annotations, is sensitive to contrast and focus drift, and relies on tiling and postprocessing that add delay.[24] AtomSegNet, trained on simulated STEM images, overcomes the scarcity of labeled experimental data and achieves high-precision atom segmentation, localization, denoising, and deblurring, but still faces challenges when experimental artifacts not captured in simulations are present.[25] Furthermore, a few-shot Prototypical Network with a ResNet101 encoder offers a path toward reduced labeling by learning from a small "support set" of labeled examples.[26] However, its core challenge for real-time use is the fundamental trade-off between adaptability and performance; while it requires fewer initial labels, its accuracy is inherently limited by the representativeness of the small support set, and it remains susceptible to failure under significant domain shift that alters feature appearance beyond the support examples.

Transfer learning[27] and foundation models[28] can also reduce labeling needs but increase compute and memory demands, require prompt or hyperparameter calibration, and may misalign with task semantics.[29] Operational reliability further depends on calibrated uncertainty, out-of-distribution detection, fail-safe behaviors, and predictable runtimes with resource-aware scheduling.

To reduce labeling demands and improve portability, domain adaptation further exploits unlabeled target data via self-supervision, few-shot learning, or template-guided[30] over-segmentation to align features with new distributions.[31] [32, 33] When gaps to prior data are large, vision foundation models (prompt-able, broadly pretrained backbones) offer zero-/few-shot starting points that can be steered or lightly adapted to new modalities. The Segment Anything Model (SAM)[28] exemplifies this approach for segmentation: trained on over a billion masks, it supports prompt-based masks across diverse imagery and can be specialized to microscopy (μSAM)[34] or guided with domain-aware post-processing (SAM-I-Am)[35] to better match scientific semantics. Because such models are already trained and efficient at inference, they are amenable to automated acquisition, where prompt-able segmentation can supply real-time control signals for closed-loop experiments. The remaining trade-offs, compute and memory footprint, prompt and hyperparameter calibration, and the need to validate domain alignment, define current best practice for deploying foundation-model–driven segmentation in scientific imaging.

SAM targets a generic "segment-everything" objective, permitting overlapping/nested masks and relying on layered crops with fixed point grids; this misaligns with microstructure semantics (regions should be exclusive and material-maximal), misses thin interfaces under sparse prompting, and over-segments homogeneous textures, all while increasing compute and latency. SAM-I-Am alleviates some errors via post-hoc "semantic boosting" (geometric filtering and texture-based mask merging), but it cannot recover regions SAM never proposed and is sensitive to thresholds, encoder quality, crop sampling, and the assumed number of materials—yielding false merges or spurious classes when textures are look-alike. Neither pipeline provides calibrated uncertainty or out-of-distribution checks to gate actions, limiting safe closed-loop use. These limitations matter for real-time, autonomous imaging and motivate methods that encode domain/physics priors, enforce exclusivity and topology, adapt prompting to feature scale, quantify uncertainty, and remain label efficient.

### I.     Physics-Aware Approach

Although a model like SAM can "segment everything," practical experiments rarely need everything segmented; they need only the physically meaningful structures that advance a specific objective. At the same time, transition to real-time data analytics requires effective hyperparameter tuning, ideally without human input. The intrinsic hyperparameters mentioned in **Table. 1** are central to this process, as they regulate how the model interprets and refines image features.

Here we introduce a physics-aware, reward-guided framework is introduced to overcome the limitations of generic segmentation by systematically embedding expert knowledge into the model hyperparameter optimization. As illustrated in **Fig. 1**, the workflow begins with image encoding, where structural information is represented as embeddings. These embeddings pass through SAM's mask and prompt decoders, producing an initial set of candidate masks. Instead of relying solely on SAM's default thresholds, this framework applies a reward layer, which scores each segmentation outcome according to physics-driven criteria such as particle overlap, morphology fidelity, or size sensitivity. The reward score is then fed back into the hyperparameter tuning loop, adjusting parameters in a systematic manner. Through repeated iterations, this closed-loop process guides the model toward segmentations that are not only computationally stable but also physically meaningful, aligning with domain-specific objectives. In this way, the segmentation task is reframed as an optimization problem, where "best" refers not to pixel-wise accuracy but to the recovery of microstructural features relevant to materials science such as grains, pores, particles, or defects—thus ensuring interpretability and scientific utility.

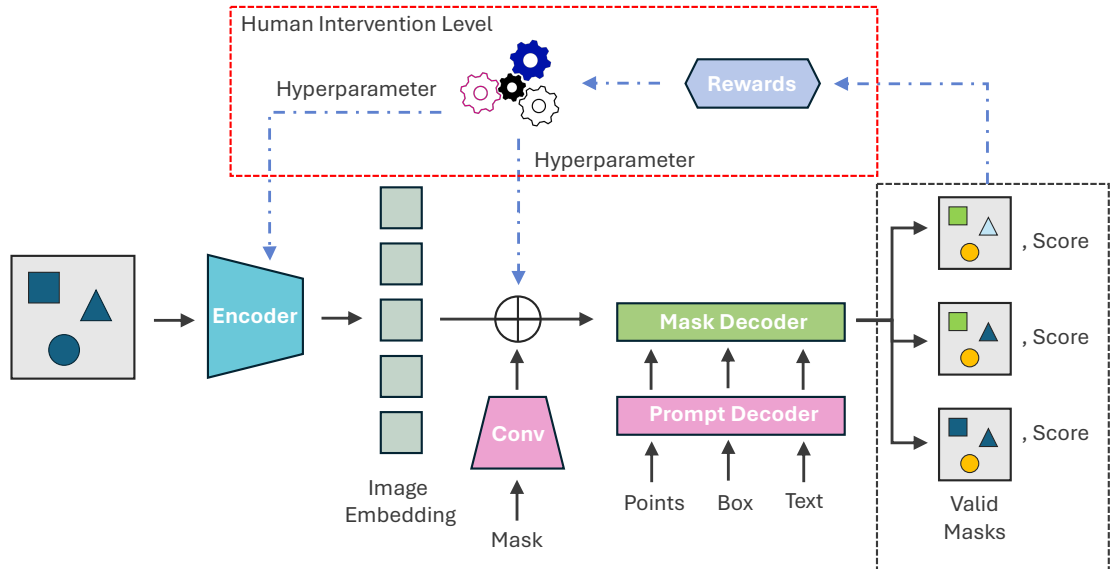

**Figure 1**: SAM* workflow

## II. Model System

We evaluate the physics-aware segmentation on three nanoparticle systems spanning complementary imaging physics—projection TEM, surface-sensitive SEM, and AFM topography (**Fig. 2**). $CsPbBr_3$ perovskite nanocrystals ($PNC$) were prepared with an $OAc: SNEA$ ligand mixture $(1:2)$ in toluene and imaged by TEM, yielding near cubic/platelet morphologies; segmentation targets single crystals and small aggregates to quantify size distributions, circularity/compactness, aspect ratio, and number density. Tin-doped indium oxide nanocrystals ($ITO\ NC$) synthesized colloidally (modified Hutchison protocol) were imaged by SEM, where segmentation isolates discrete particles and agglomerates to extract projected area, shape anisotropy (aspect ratio, convexity/solidity), inter-particle spacing, and areal coverage. Thermally dewet gold–cobalt thin film nanoparticles ($AuCo\ NP$) deposited on a $SiO_2$ substrate was characterized by AFM, with segmentation operating on height maps to delineate island footprints and compute footprint area, particles size, and surface coverage. Together these materials provide a concise, materials-grounded testbed for assessing task-aligned, physics-aware segmentation.

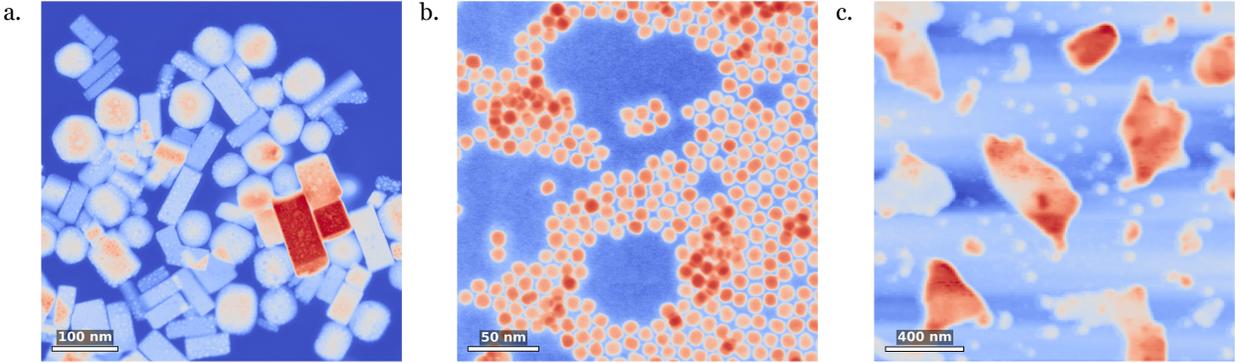

**Figure 2**: Representative datasets: (a) HAADF-STEM of $CsPbBr_3$ perovskite nanocrystals; (b) SEM of indium–tin oxide ($ITO$) nanocrystals; (c) AFM topography of $Au-Co$ nanoparticles on $SiO_2$. False-color renderings for clarity. Scale bars: 100 nm (a), 50 nm (b), 400 nm (c).

III. **Operational Status of SAM Hyperparameters**

Vanilla SAM, implemented as the AutoMaskGenerator pipeline, generates and filters segmentation masks in the absence of task-specific priors. The framework is composed of three core modules: an image encoder, a prompt encoder, and a lightweight mask decoder. The image encoder is based on a Vision Transformer (ViT)[36] pre-trained with Masked Autoencoders (MAE)[37] and adapted for high-resolution inputs. It processes an image once to produce a high-dimensional embedding, a computationally expensive step that can be reused for subsequent prompts. The prompt encoder accepts both sparse and dense inputs. Sparse prompts including points, bounding boxes, and free-form text are represented by positional encodings combined with learned embeddings, while text prompts are processed using a pre-trained CLIP encoder.[38] Dense prompts in the form of masks are embedded with convolutional layers and added element-wise to the image embedding. The mask decoder then integrates the image embedding, prompt embeddings, and an output token to generate segmentation masks. It employs a modified Transformer decoder with prompt self-attention and bidirectional cross-attention between image and prompt features. After two decoding blocks, the embedding is up-sampled, and a multi-layer perceptron (MLP)[39] transforms the output token into a dynamic classifier that estimates the foreground probability across the image. A defining feature of SAM is its ambiguity awareness: for uncertain prompts (e.g., a point that could denote a shirt or the person wearing it), the model outputs multiple valid masks (typically three) each assigned a confidence score based on predicted Intersection over Union (IoU). Training is carried out using a promptable segmentation task, with geometric prompts simulated to mimic interactive use, and optimized via focal and Dice loss functions.

In practice, the behavior of this pipeline is governed by a compact set of hyperparameters (**Table 1**) that regulate (i) search density: the spatial density of automatic prompts and optional multi-scale crops, (ii) proposal quality thresholds: gates on predicted Intersection over Union (IoU) and stability, (iii) deduplication: non-maximum suppression across boxes and crops, and (iv) an explicit area filter on connected components. For a fixed image, these controls jointly determine the operating point in terms of recall, precision, size selectivity, and runtime. When the pixel scale changes (e.g., different magnification or resizing), only the size-dependent terms (most notably the area cutoff) require rescaling; the remaining thresholds and search parameters can typically be held constant or tuned within narrow bands.

**Table 1**: AutoMaskGenerator hyperparameters with operational status

| | Parameter | Role | Span | Primary effect | Operational status |
|---|---|---|---|---|---|
| 1 | min_mask_region_area (px²) | Hard area gate on connected components | Set from physics† | Removes masks below cutoff (↑ size selectivity; ↓ small-object recall) | Scenario-dependent |
| 2 | pred_iou_thresh | pred. IoU | 0.85–0.99 | ↑ precision; prunes weak/partial masks; may drop faint/small | Scenario-dependent |
| 3 | stability_score_thresh | Gate on robustness to prompt perturbations | 0.90–0.99 | ↓ unstable/jagged masks; too high can remove thin structures | Scenario-dependent |
| 4 | stability_score_offset | Internal normalization for stability score | 1.0 (fix) | Minimal direct effect | Fixed |
| 5 | box_nms_thresh | NMS IoU among proposals within a crop/scale | 0.20–0.70 | Lower → more suppression (fewer duplicates); higher → more overlaps | Scenario-dependent |
| 6 | crop_nms_thresh | NMS IoU across overlapping crops | 0.20–0.70 | Lower → stronger cross-crop de-dup; higher → retain overlaps | Effectively fixed if crop_n_layers=0; otherwise, scenario-dependent |
| 7 | points_per_batch | Batching for throughput/memory | 32–128 | Affects speed/memory only; no change to masks | Fixed (set once for hardware) |
| 8 | output_mode | Output encoding (binary/RLE) | — | No effect on masks; serialization only | Fixed |
| 9 | points_per_side | Auto-prompt grid density per image/crop | 6–64 (even ints) | ↑ proposals & small-object recall; ↑ runtime/duplicates | Scenario-dependent |
| 10 | crop_n_layers | # of multi-scale crops (pyramid levels) | 0–2 (occasionally 3) | ↑ scale coverage (very small/large); ↑ compute & duplicates | Scenario-dependent |
| 11 | crop_overlap_ratio | Fractional overlap between neighboring crops | 0.20–0.60 | ↑ boundary recovery; ↑ duplicates/compute | Effectively fixed if crop_n_layers=0; otherwise, scenario-dependent |
| 12 | crop_n_points_downscale_factor | Prompt-grid thinning on deeper crop layers | 1–4 | >1 makes prompts sparser at coarse scales (↓ compute; ↓ small recall there) | Effectively fixed if crop_n_layers=0; otherwise, scenario-dependent |
| 13 | point_grids | Custom prompt layouts | — | Targets bespoke spatial sampling | Effectively fixed if "None"; otherwise, scenario-dependent |

Across different images, the operating point of Vanilla SAM shifts because several hyperparameters are scale-variant while others are largely scale-invariant. Changes in pixel scale (magnification, binning, or resizing) and object size distribution directly affect search density and the area gate. The effective prompt spacing of points_per_side (pps) becomes coarser on larger images and denser on smaller ones, and min_mask_region_area (mmra) must be recomputed in pixels from the desired physical cutoff (minimum particle diameter). In contrast, quality thresholds (pred_iou_thresh (pit), stability_score_thresh (sst)) and deduplication settings (box_nms_thresh (bnt), crop_nms_thresh (cnt)) can usually be held within narrow bands and only adjusted for image quality and scene density (lower thresholds for noisy/low-contrast data to preserve recall; stronger Non-Maximum Suppression (NMS) for crowded scenes to suppress duplicates).

IV. Single-Objective Framework (Overlapping Scenario)

We illustrate the hyperparameter sensitivity of SAM for the Indium–tin oxide nanocrystals ($ITO\ NC$). Here, the vanilla SAM configuration frequently under-segments particles that partially occlude one another, as illustrated in **Fig. 3.** In such dense particle ensembles, segmentation fidelity is governed by a large set of hyperparameters that directly modulate the model's sensitivity to overlap. The points_per_side determines the sampling density within each crop; increasing this density enhances the likelihood of probing inter-particle boundaries, which is essential for separating partially occluded nanocrystals. The stability_score_thresh regulates mask acceptance based on contour consistency. While high thresholds suppress noise, they also tend to reject masks in regions of overlap where boundaries are less stable, thereby reducing overlap sensitivity. The box_nms_thresh and crop_nms_thresh set the tolerance of non-maximum suppression within and across crops; conservative values merge adjacent detections into a single mask, whereas relaxed thresholds preserve distinct boundaries between overlapping particles. The crop_overlap_ratio further amplifies overlap recovery by introducing redundancy across neighboring crops, ensuring that particles straddling crop borders (and often overlapping) are repeatedly proposed. Finally, crop_n_layers is critical: in vanilla SAM this parameter is set to zero, meaning no hierarchical cropping is performed and overlapping particles are systematically missed; enabling nonzero layers introduces multi-scale, overlapping crops that substantially improve the detection of occluded particles.

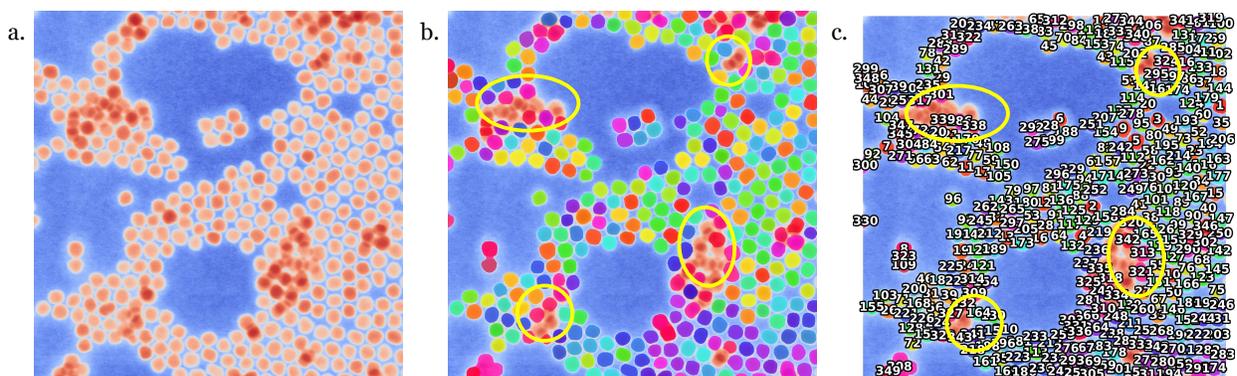

Figure **3**: Vanilla SAM segmentation. (a) Input image. (b) Automatic mask proposals from SAM (no input of prior knowledge or per-post-processing); (c), regions are uniquely color-coded and indexed. 349 masks were detected, overlapping particles is missed segmented.

These hyperparameters jointly orchestrate a delicate balance between sensitivity and specificity. When tuned appropriately, they enable SAM to recover overlapping nanocrystals while minimizing false positives and merged artifacts. Without such tuning, as shown in **Fig. 3**, vanilla SAM produces 349 masks yet consistently fails to segment overlapping particles.

A geometry-aware reward formulation provides a principled framework for systematically tuning SAM hyperparameters to enhance the recovery of overlapping particles while suppressing artifacts. Consider a predicted mask set $\mathcal{M} = \{m_1, m_2, \ldots, m_N\}$. For each mask pair $(m_i, m_j)$, we compute the intersection-over-union ($IoU$). From this, we define three key quantities:

$$O(\mathcal{M}) = |\{(m_i, m_j): \tau_l \leq IOU(m_i, m_j) \leq \tau_h, i \neq j\}| \tag{1}$$

$$D(\mathcal{M}) = |\{(m_i, m_j): IOU(m_i, m_j) \geq \tau_{dup}, i \neq j\}| \tag{2}$$

$$B(\mathcal{M}) = |\{m_i \in \mathcal{M}: A(m_i) \geq \gamma \cdot median_j(A(m_j))\}| \tag{3}$$

Where, $O(\mathcal{M})$ is the number of distinct but partially overlapping pairs of masks. The thresholds $\tau_l$ and $\tau_h$ define the $IoU$ range for valid overlaps ($0.1 \leq IoU \leq 0.60$). $D(\mathcal{M})$ is the number of duplicate pairs, where masks overlap almost completely ($IoU \geq \tau_{dup}$ e.g., 0.9). $B(\mathcal{M})$ is the count of merged candidates, defined as masks whose area $A(m_i)$ exceeds γ times the median mask area (e.g., $\gamma = 3$). With these definitions, the reward function is:

$$\mathcal{F}(\mathcal{M}) = \frac{1 + O(\mathcal{M})}{1 + \alpha \cdot D(\mathcal{M}) + \beta \cdot B(\mathcal{M})} \tag{4}$$

where α and β weight the penalties for duplicates and merged masks. This formulation rewards the presence of distinct overlapping particles $O(\mathcal{M})$, while penalizing both redundant detections $D(\mathcal{M})$ and spurious merges $B(\mathcal{M})$. By balancing these contributions, the reward biases segmentation toward physically meaningful particle ensembles in crowded, overlapping regimes.

With physics-aware tuning (SAM*), segmentation performance improves substantially compared to the vanilla configuration. A total of 415 masks were detected, exceeding the 349 obtained previously, indicating enhanced sensitivity in crowded regions. Critically, particles that overlap with one another (highlighted in the circled regions) are now segmented as distinct entities rather than being merged or missed. This improvement is attributable to adjustments in non-maximum suppression and crop redundancy, which allow partially overlapping proposals to survive the filtering stage. As shown in **Fig. 4**, each particle is uniquely color-coded and indexed, confirming that SAM* not only increases detection count but also preserves the individuality of nanocrystals within dense clusters.

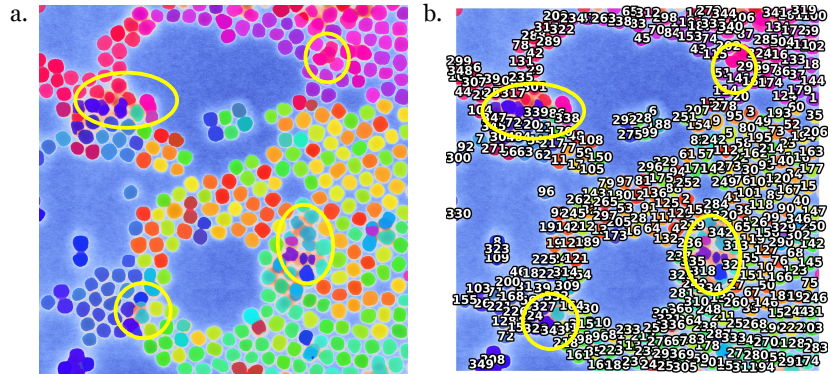

*Figure 4*: SAM* segmentation. (a) mask proposals from SAM* (with input of prior knowledge); regions are uniquely color-coded and indexed. 415 masks were detected, overlapping particles are segmented.

## V. Multi-Objective Framework (Clustering Scenario)

A distinctive feature of the $CsPbBr_3$ perovskite nanocrystal ensemble is the coexistence of multiple particle morphologies (nearly circular nanocrystals intermixed with elongated, rectangular crystallites as presented in **Fig. 5a**. Beyond simple detection, meaningful segmentation in this context requires morphology awareness, ensuring that circular and rectangular particles are not conflated into a single undifferentiated set of masks. Such discrimination is essential for downstream clustering, where particle shape serves as a key descriptor of growth pathways, surface energetics, and phase-dependent behavior. Vanilla SAM is unable to differentiate between circular and rectangular nanocrystals because its segmentation process is fundamentally geometry-agnostic. The algorithm is designed to propose stable, closed contours without embedding explicit priors about particle shape, meaning that a circle and a rectangle of similar area can be segmented with equally high confidence.

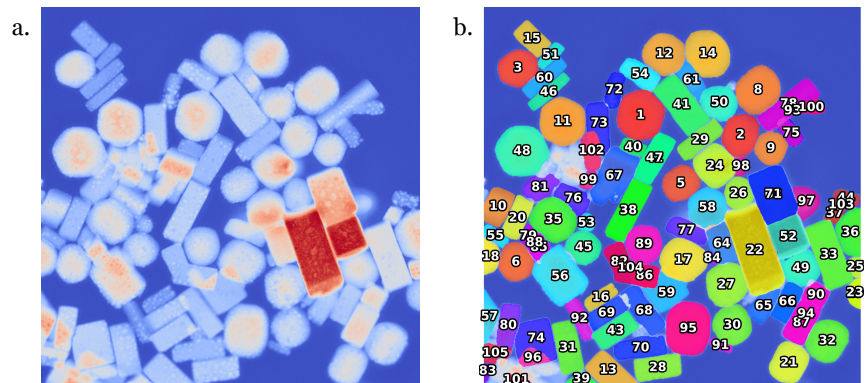

**Figure 5**: Vanilla SAM segmentation. (a) Input image. (b) Automatic mask proposals from SAM (no input of prior knowledge or per-post-processing); regions are uniquely color-coded and indexed. 105 masks were detected, overlapping particles is missed segmented, also no separation between particle morphology.

This example presents a considerably more complex challenge than simple adjustment of SAM's hyperparameters. Hyperparameter tuning can bias the sensitivity of the segmentation pipeline. For example, lowering thresholds may allow elongated, low-stability masks to be retained, while raising them may preferentially suppress small, circular ones. Such adjustments, however, merely influence which candidate masks survive the filtering process; they do not endow SAM with an explicit capacity to discriminate geometric classes. In other words, while tuning may shift the balance toward detecting more rectangular or more circular masks depending on their response to stability, $IoU$, and non-maximum suppression criteria, the model itself lacks any intrinsic notion of "rectangularity" or "circularity."

The core idea to overcome SAM's inherent geometry-agnostic nature is to define shape-sensitive reward functions that guide the hyperparameter search process toward segmentation outcomes enriched in either circular or rectangular nanocrystals. For a given hyperparameter configuration $h$, SAM generates a mask set $\mathcal{M}(h) = \{m_1, m_2, \ldots, m_N\}$, where each mask $m_i$ is characterized by descriptors including area $A(m_i)$, perimeter $P(m_i)$, and bounding box dimensions $(\omega_i, h_i)$. From these descriptors, two morphology indicators are derived:

$$\mathcal{F}_1(m_i) = \frac{4\pi A(m_i)}{P(m_i)^2} \qquad [5]$$

$$\mathcal{F}_2(m_i) = max(\frac{\omega_i}{h_i}, \frac{h_i}{\omega_i}) \qquad [6]$$

Where, $\mathcal{F}_1(m_i)$ quantifies the circularity, approaching unity for perfect disks and decreasing for elongated shapes, whereas $\mathcal{F}_2(m_i)$ captures the aspect ratio, with values near one for isotropic particles and larger values for rod-like or rectangular morphologies. The segmentation task is then posed as a multi-objective optimization problem: maximize $\mathcal{F}_1(\mathcal{M})$ and $\mathcal{F}_2(\mathcal{M})$ simultaneously, subject to penalties for duplicate or merged masks. Within this framework, different hyperparameter configurations $h \, \epsilon \, H$ naturally bias the segmentation process toward one or the other objective. For example, one region of the hyperparameter space may enrich circular detections at the expense of rods, while another favors rectangular detections but suppresses disks. By evaluating candidate configurations against both objectives, one obtains a spectrum of morphology-aware trade-offs, each corresponding to a distinct balance between circular and rectangular particle recovery.

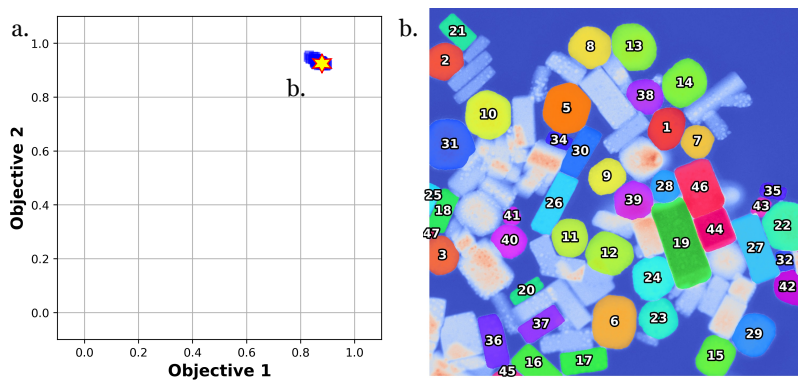

**Figure 6**: Multi-objective tuning of SAM for size-selective segmentation (SAM*). (a) Pareto front in the Objective-1/Objective-2 planes; each point is a distinct SAM* hyperparameter vector. (b) Trade-off operating point from the Pareto set: 47 instances masks were detected.

In **Fig. 6**, vanilla SAM was tuned in an attempt to introduce morphology awareness into the framework by adjusting its hyperparameters. The Pareto front solutions in **Fig. 6a**, where each point represents a distinct combination of hyperparameters associated with a particular reward outcome, did not produce a broad spread of trade-offs. Instead, the points collapsed into a narrow cluster. This behavior arises because the tunable hyperparameters (such as points_per_side, pred_iou_thresh, and box_nms_thresh) primarily regulate the number of masks, their stability, and redundancy, but do not encode any intrinsic geometric preference. As a result, both circularity and aspect-ratio–based objectives shifted in tandem rather than diverging, leading the rewards to accumulate or "compile" together rather than exhibiting antagonistic trade-offs. While SAM* was able to generate consistent mask boundaries, it failed to discriminate between circular and rectangular nanocrystals, underscoring a fundamental limitation: vanilla SAM is geometry-agnostic. Hyperparameter tuning alone cannot impart morphology awareness and therefore cannot selectively enrich detections of specific shapes such as disks versus rods. Achieving genuine morphology-sensitive segmentation would require augmenting the framework with shape-sensitive layers or losses directly into SAM's training pipeline. Nonetheless, it is important to recognize that once segmentation masks are obtained, classification into circular or rectangular populations is straightforward using simple post-processing descriptors such as circularity or aspect ratio.

VI. **Multi-Objective Framework (Antagonistic Rewards Scenario)**

In microscopy of supported nanomaterials, it is often the smallest features that govern catalytic activity, stability, and coarsening pathways, yet they are also the first to escape detection when generic segmentation tools are applied. The example of such problem is the $AuCo$ nanoparticle **Fig. 7a**, where the challenge lies in capturing both isolated nanoparticles and extended aggregates within the same field of view. When the vanilla SAM pipeline is applied, without any task-specific priors or physics-aware calibration, it produces 79 candidate masks **Fig. 7b**, each color-coded and indexed as an independent segmentation proposal.

Although this outputs underscores SAM's generic capacity for region proposal, several systematic deficiencies emerge in the context of materials imaging. First, some nanoparticles below ≈100 nm are absent from the detection set, reflecting a pronounced recall bias toward mesoscale aggregates. Second, extended domains are frequently over-partitioned, producing multiple discrete masks for physically contiguous structures and thereby breaking physical connectivity that is known to exist in the sample. Third, the overall mask ensemble is statistically unbalanced, containing spurious fragments while under-representing the nanoscale particle population that is central to quantitative analysis.

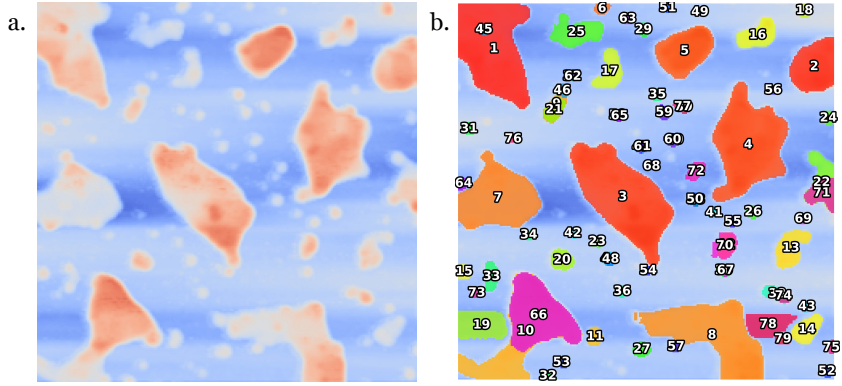

**Figure 7**: Vanilla SAM segmentation. (a) Input image. (b) Automatic mask proposals from SAM (no input of prior knowledge or per-post-processing); regions are uniquely color-coded and indexed. 79 masks were detected.

In this case of, two competing physical criteria define the task: (i) maximizing sensitivity to small particles and (ii) maximizing coverage of large aggregates. To capture these objectives, we define reward functions directly on mask area distributions produced by SAM. For small-particle sensitivity, the reward is designed to favor detection of minimal regions by penalizing mask area:

$$\mathcal{F}_1(\mathcal{M}) = \frac{1}{|\mathcal{M}|} \sum_{m \in \mathcal{M}} \frac{1}{A(m) + \varepsilon} \qquad [7]$$

where $\mathcal{M}$ is the set of masks, $A(m)$ the pixel area of mask $m$, and $\varepsilon$ a small constant to avoid division by zero. This reward increases when SAM identifies many small, valid objects.

Conversely, for large-particle coverage, we maximize the mean detected area:

$$\mathcal{F}_2(\mathcal{M}) = \frac{1}{|\mathcal{M}|} \sum_{m \in \mathcal{M}} A(m) \qquad [8]$$

which promotes masks that cover extended structures. By jointly optimizing $\mathcal{F}_1$ and $\mathcal{F}_2$, we construct a multi-objective reward landscape in which hyperparameters such as points_per_side (pps), points_per_batch (ppb), and pred_iou_thresh (pit) are tuned to reconcile two competing aims: (i) fine-grained discovery of nanoscale particles, and (ii) robust coverage of extended aggregates. Rather than relying on ad-hoc visual plausibility, this framework anchors segmentation quality in quantitative, material-specific criteria.

In this setting, Pareto optimization naturally yields trade-offs that are *physics-aligned*, producing parameter configurations tailored to the experimental system at hand. The result is a transformation of SAM, from a broad "segment anything" tool, into a "smart segmentation" engine that learns to segment what matters.

To embed physical criteria into segmentation, SAM* is optimized using a multi-objective reward formulation in which hyperparameter vectors are evaluated against two competing objectives: $\mathcal{F}_1$, which rewards large, contiguous masks representing extended aggregates, and $\mathcal{F}_2$, which rewards the detection of nanoscale particles by emphasizing smaller mask areas. The resulting optimization space forms a Pareto front in the $\mathcal{F}_1$–$\mathcal{F}_2$ plane illustrated in **Fig. 8a**, where each point denotes a distinct SAM* configuration that cannot be simultaneously improved in both criteria. This landscape encodes the trade-off between fine-grained small-particle recovery and faithful representation of extended structures. A configuration biased toward $\mathcal{F}_1$–$\mathcal{F}_2$ as shown in **Fig. 8b** yields 117 detected instances, substantially improving the recall of sub-100 nm particles compared to vanilla SAM. However, this gain comes at the expense of over-fragmenting large islands into multiple disconnected sub-masks, distorting the physical integrity of extended features. At the opposite extreme, an $\mathcal{F}_1$–$\mathcal{F}_2$-oriented configuration presented in **Fig. 8d** produces only 11 masks, each corresponding to large aggregates represented as coherent structures. In this case, nearly all nanoparticles are excluded, reflecting a strong bias toward maximizing large-particle coverage at the cost of small-particle sensitivity. Between these two extremes, a Pareto-balanced configuration **Fig. 8c** identifies 22 instances that better reconcile the competing objectives. In this regime, both nanoscale particles and larger islands are represented, producing a more uniform distribution of detections across size scales. The outcome demonstrates that multi-objective tuning transforms SAM from a generic "segment anything" model into SAM*, a physics-aware segmentation engine where masks are evaluated not by visual plausibility alone but by their relevance to experimental size criteria.

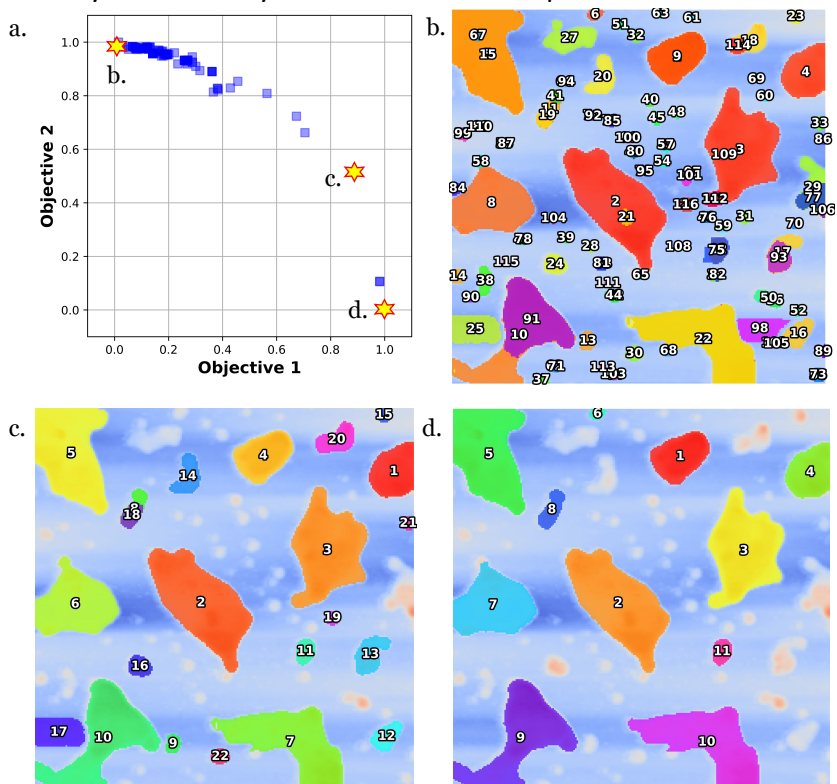

**Figure 8**: Multi-objective tuning of SAM for size-selective segmentation (SAM*). (a) Pareto front in the Objective-1/Objective-2 planes; each point is a distinct SAM* hyperparameter vector. (b) Configuration biased toward Objective 2 (small-particle recall): 117 instances detected, recovering many sub-100-nm particles. (c) Trade-off operating point from the Pareto set: 22 instances with more uniform performance across size scales. (d) Configuration biased toward Objective 1 (large-particle recall): 11 instances, capturing large islands as contiguous masks with few small detections.

**Conclusion**

Segmentation in microscopy has long been constrained by a tension between general-purpose algorithms and the specificity demanded by scientific inquiry. Foundation models such as SAM offer broad segmentation capability but still require multiple hyperparameters required to tune the segmentation process. For imaging data emerging in the context of the microscopy, in their default form they systematically miss structures of greatest experimental relevance such as overlapping nanocrystals that are merged or discarded, morphological classes that remain conflated, and nanoscale particles that disappear amidst aggregates. These limitations arise because the model is optimized for generic visual plausibility rather than for physics-aware fidelity.

By embedding reward-driven optimization directly into the segmentation pipeline, SAM is reoriented from a geometry-agnostic foundation model into SAM*, allowing the unsupervised hyperparameter tuning for specific applications. The introduction of quantitative reward functions fundamentally reframes segmentation: overlap fidelity and size sensitivity become explicit optimization objectives rather than incidental outcomes of heuristic thresholding. However, because SAM itself is not morphology-aware, SAM* cannot intrinsically classify circular versus rectangular particles. This would require incorporating shape-sensitive priors or additional learning layers beyond hyperparameter tuning. Applied across diverse nanomaterial systems, this approach nonetheless enabled recovery of occluded ITO nanocrystals and reconciliation of antagonistic objectives in AuCo ensembles where sub-100 nm particles coexist with extended aggregates. In each case, SAM* improved detection and preserved physical interpretability, ensuring computational results remained aligned with experimental reality.

More broadly, the reward-guided framework elevates segmentation from a static, general-purpose routine to a dynamic, task-adaptive methodology. Its utility extends well beyond nanomaterials: any domain in which the value of segmentation lies in capturing meaningful structural relationships can benefit from embedding expert knowledge as quantitative objectives, where the definition of 'best' segmentation is grounded not in abstract pixel-wise accuracy but in its ability to extract physically meaningful insight.

**SUPPLEMENTARY INFORMATION**

The Supplementary Information details the optimization of SAM hyperparameters using the NSGA-II Genetic Algorithm,[40] addressing three scenarios: (i) single-objective tuning for overlapping particles, (ii) multi-objective clustering with morphology-sensitive rewards, and (iii) multi-objective trade-offs between nanoscale particle detection and aggregate coverage.

**AUTHOR DECLARATIONS**
Conflict of Interest: The authors have no conflicts to disclose.

**Author Contributions:**
Kamyar Barakati: Conceptualization (equal), Data curation (equal), Formal analysis (equal), Writing – original draft (equal), Methodology (equal); Sergei V. Kalinin: Conceptualization (equal), Formal analysis (equal), Funding acquisition (equal), Writing – review & editing (equal), Supervision (equal); Aditya Raghavan: Investigation (equal); Utkarsh Pratiush: Software (equal), Investigation (equal); Sheryl L. Sanchez, Delia J. Milliron, Mahshid Ahmadi, Philip D. Rack: Resources (equal).

**DATA AVAILABILITY:**

The data that support the findings of this study are available from the corresponding authors upon request.

**ACKNOWLEGEMENTS:**

This work (workflow development, reward-driven concept) was supported (K.B., S.V.K., and A.R.) by the U.S. Department of Energy, Office of Science, Office of Basic Energy Sciences as part of the Energy Frontier Research Centers program: CSSAS-The Center for the Science of Synthesis Across Scales under award number DE-SC0019288. The work was partially supported (U.P.) by AI Tennessee Initiative at University of Tennessee Knoxville (UTK). This research (synthesis of the Au-Co thin film alloy) was partially supported (P.D.R.) by the National Science Foundation Materials Research Science and Engineering Center program through the UT Knoxville Center for Advanced Materials and Manufacturing (DMR-2309083). (S.S. and M.A.) acknowledge support from the National Science Foundation (NSF), Award Number 2043205. (S.S.) acknowledge partial support from the Center for Materials Processing (CMP) at the University of Tennessee, Knoxville. (D.J.M.) acknowledges support from the National Science Foundation (NSF), Award Number CHE-2303296.